\documentclass[11pt]{article}
\pdfoutput=1
\usepackage{etex}
\usepackage{microtype} %
\usepackage{booktabs}  %
\usepackage{url}  %
\usepackage{siunitx}
\usepackage{graphicx}
\usepackage[flushleft]{threeparttable} %
\usepackage{xcolor}

\usepackage{amsmath}
\usepackage{amsthm}

\usepackage[finalworkshop]{automl}

\usepackage{natbib}
\title{Assembled-OpenML: Creating Efficient Benchmarks for Ensembles in AutoML with OpenML}

\author[1]{\nameemail{Lennart Purucker}{lennart.purucker@uni-siegen.de}}
\author[1]{\nameemail{Joeran Beel}{joeran.beel@uni-siegen.de}}

\affil[1]{University of Siegen}

\hypersetup{%
  pdfauthor={Purucker, Lennart and Beel, Joeran}, %
  pdftitle={Assembled-OpenML: Creating Efficient Benchmarks for Ensembles in AutoML with OpenML},
  pdfsubject={A tool to collect predictions of runs from OpenML to be used in ensemble techniques.},
  pdfkeywords={AutoML, Ensemble, Stacking, Dynamic Selection, Benchmark, Meta-Datasets}
}

\begin{document}
\maketitle

\begin{abstract}
Automated Machine Learning (AutoML) frameworks regularly use ensembles.
Developers need to compare different ensemble techniques to select appropriate techniques for an AutoML framework from the many potential techniques. 
So far, the comparison of ensemble techniques is often computationally expensive, because many base models must be trained and evaluated one or multiple times.
Therefore, we present Assembled-OpenML. Assembled-OpenML is a Python tool, which builds meta-datasets for ensembles using OpenML.
A meta-dataset, called Metatask, consists of the data of an OpenML task, the task's dataset, and prediction data from model evaluations for the task. 
We can make the comparison of ensemble techniques computationally cheaper by using the predictions stored in a metatask instead of training and evaluating base models.
To introduce Assembled-OpenML, we describe the first version of our tool. Moreover, we present an example of using Assembled-OpenML to compare a set of ensemble techniques. 
For this example comparison, we built a benchmark using Assembled-OpenML and implemented ensemble techniques expecting predictions instead of base models as input.
In our example comparison, we gathered the prediction data of $1523$ base models for $31$ datasets. Obtaining the prediction data for all base models using Assembled-OpenML took ${\sim} 1$ hour in total. In comparison, obtaining the prediction data by training and evaluating just one base model on the most computationally expensive dataset took ${\sim} 37$ minutes.  
\end{abstract}

\section{Introduction}
Combining the predictions of several models can produce better overall predictions \citep{dietterich2000ensemble, kittler1998combining}.
Building an ensemble from a set of base models is a common practice in Machine Learning (ML) and Automated Machine Learning (AutoML).
Real-world ML applications regularly use ensembles, see \citep{hakak2021ensemble, gunturi2021ensemble, chung2019ensemble}. 
AutoML frameworks often have ensembles as just another algorithm in the search space during model selection (e.g., Random Forest, XGBoost, etc.). 
Alternatively, frameworks treat ensembles as a special part of the search space and focus on building one large ensemble. For example, AutoGluon \citep{erickson2020autogluon}, Autostacker \citep{chenAutostackerCompositionalEvolutionary2018}, and Automatic Frankensteining \citep{wistubaAutomaticFrankensteiningCreating2017} use stacked generalization \citep{DBLP:journals/nn/Wolpert92} and explicitly focus on building an ensemble.
Lastly, ensembles can be built post hoc by using (a subset of) all models found during model selection. 
Here, AutoGluon and Auto-Sklearn \citep{feurer-neurips15a} use ensemble selection from libraries of models \citep{caruana2004ensemble}. 
Post hoc ensembling can be formulated as its own optimization problem, and initial work in this direction has already been done \citep{zhao2022autodes}.

Many different ensemble techniques could be used in AutoML. Techniques from related fields such as dynamic ensemble or classifier selection could also be used \citep{ko2008dynamic, giacinto2001dynamic}. 
Hence, ensemble techniques must be compared to find the techniques that are best as part of the search space, as a special part of the search space, or for post hoc ensembling.

However, comparing ensemble techniques is computationally expensive, and there are no dedicated benchmarks for ensemble techniques to speed up the comparison.
Usually, new and existing techniques are compared by evaluating them on ML datasets. Moreover, the ensembles are built from a potentially large set of pre-selected base models. 
Such comparisons are unnecessarily computationally expensive, because base models are trained and evaluated one or multiple times for every comparison. 
To illustrate, base models are commonly trained and evaluated for every dataset once or each ensemble technique individually trains and evaluates the base models when needed; see \citep{cruz2018dynamic, zhao2008enhanced, cid2015discrete, van2018online, alvarez2015classifier}.

We considered two solutions to enable less computationally expensive comparisons by avoiding the computational overhead of base models. 
First, compute and share trained base models for a benchmark set of datasets. By building ensembles from pre-trained base models, we can avoid the cost of training base models for every comparison. 
Second, compute and share the predictions of trained base models. Thus, allowing us to avoid training and evaluating base models by building ensembles from the prediction data instead of models. This is possible since we only require the base models' predictions to build and evaluate most ensemble techniques. 

In this paper, we introduce a tool for the second solution, dubbed \emph{Assembled-OpenML}.
As the name suggested, we build upon the OpenML platform \citep{DBLP:journals/corr/VanschorenRBT14} and its ecosystem of tools\footnote{See \url{https://github.com/openml} for OpenML's library of tools.}, which enable ML users to share and reuse machine learning (meta-)data. 
In detail, we present a Python tool that can automatically build a set of meta-datasets, called \emph{Metatasks}, using data from OpenML.
Assembled-OpenML selects a set of base models for an ML task from OpenML to create metatasks. %
A metatask contains data on the original task, the associated dataset, additional metadata, and each selected base model's predictions as well as confidences. All required (meta-)data for a metatask are fetched from OpenML. 

Metatasks can be used to compare ensemble techniques while being less computationally expensive. 
The data stored in a metatasks allows us to simulate ensemble techniques, that is, execute ensemble techniques without having to train and evaluate base models. Thus, leaving only the computational overhead of the ensemble techniques.
To illustrate, we were able to build $31$ metatasks containing the prediction data of $1523$ base models in ${\sim} 1$ hour using Assembled-OpenML. Training and evaluating just one base model to obtain its prediction data on the most computationally expensive dataset took ${\sim} 37$ minutes.  

In this paper, we present the first version of Assembled-OpenML.
Our contribution with Assembled-OpenML is the automation of creating metatasks and thus enabling efficient benchmarks for ensembles.
Moreover, we show its use by simulating and comparing $5$ ensemble techniques and $3$ baselines on an example benchmark set of $31$ metatasks. 
The framework Assembled-OpenML, example usage code, and data related to this paper can be found in our GitHub repository\footnote{\url{https://github.com/ISG-Siegen/assembled}}.

\section{Related Work}
In related research fields, like hyperparameter optimization and neural architecture search, the computational cost of comparing optimization techniques motivated the creation of surrogate and tabular benchmarks.  
Surrogate benchmarks provide a surrogate model that is used to predict the performance of a configuration such that the expensive evaluation of the configuration can be avoided \citep{DBLP:conf/aaai/EggenspergerHHL15}. 
Tabular benchmarks provide a look-up table to obtain the performance of configurations \citep{DBLP:conf/icml/YingKCR0H19,klein2019tabular}.

Both types of benchmarks do not exist for ensemble techniques. Moreover, to the best of our knowledge, no previous work has tried to reduce the computational cost of comparing ensembles. We do not know of any work that systematically created, stored, and shared the predictions of base models. 
Likewise, we do not know of any appropriate repository of trained base models for ensembles. 
The closest to this would be the model zoo\footnote{\url{https://github.com/BVLC/caffe/wiki/Model-Zoo}} from Caffe \citep{jia2014caffe} or the model garden\footnote{\url{https://github.com/tensorflow/models}} from TensorFlow \citep{tensorflowmodelgarden2020} for transfer learning with pre-training. Both store models of deep neural networks for computer vision, natural language processing, or recommender systems.
However, they are not appropriate because they do not store trained (traditional) models for tabular classification or regression tasks.

OpenML was frequently used to produce meta-datasets. So far, OpenML has been used mainly to produce meta-datasets consisting of meta-features in terms of complexity measures (like the number of instances of a dataset) and the performance of algorithms for a specific metric \citep{bilalli2017predictive, tornede2020extreme, olier2018transformative, kuhn2018automatic}. Such undertakings also produced a set of basic tools to extract meta-data from OpenML\footnote{For example: \url{https://github.com/joaquinvanschoren/openml-metadata/} or \url{https://github.com/bbilalli/MetadataFromOpenML}}.
Yet, none of these undertakings extracted predictions from OpenML.

Lastly, we are not aware of any exhaustive comparison or benchmark of ensemble techniques. The closest we have found to this is a GitHub repository with a Per Instance Algorithm Selection Benchmark for Multi-Classifier Systems and AutoML by Edward Bergman\footnote{\url{https://github.com/eddiebergman/piasbenchmark}}. This benchmark could be reused or extended to compute and share trained models instead of predictions using OpenML. 

\section{Assembled-OpenML: a Meta-Dataset Collection Framework}
\label{sec:Assembled-OpenML}
Assembled-OpenML expects a task ID from OpenML as input to build a metatask. For reference, ${\sim} 4100$ classification and ${\sim} 19000$ regression tasks exist on OpenML at the time of writing. 
First, Assembled-OpenML fetches the original OpenML task. The OpenML task is used to collect the dataset and all information related to the dataset (e.g., a predefined validation split). 
Next, Assembled-OpenML fetches the set of top-n best performing configurations of the task according to a user-selected metric.  
Thereby, Assembled-OpenML ensures that the top-n set does not contain duplicated configurations. 
Lastly, Assembled-OpenML parses and collects the prediction data of each configuration in the top-n set.
The prediction data includes the predictions and their confidences. 
To clarify a relevant edge case, we store the concatenated prediction data of each fold if an OpenML task used cross-validation. 
We use the Python extension for OpenML by \cite{feurer-arxiv19a} to fetch data from OpenML\footnote{See Appendix \ref{appdx:usedtools} for the references, versions, and licences of all required Python libraries.}.

In this initial version, Assembled-OpenML still faces some limitations. 
Assembled-OpenML only supports classification tasks so far. 
The number of OpenML runs for a task represents the available amount of prediction data. 
Unfortunately, the number of runs available on OpenML for regression tasks is very small compared to the number of runs for classification tasks. To illustrate, the top 10 classification tasks have between ${\sim}417000$ and ${\sim}159000$ runs at the moment, while the top 10 regression tasks have between ${\sim}160$ and $18$ runs. 
Therefore, we focused on classification for the initial version. 
Similarly, we ignore evaluation repetitions because the vast majority of tasks do not include repetitions. 
Lastly, we encountered problems with corrupted prediction data\footnote{See Appendix \ref{appdx:parseproblems} for more details on corrupted prediction data.}. 

Assembled-OpenML does not put any constraints on the top-n set besides ignoring duplicates. Additional constraints might not be appropriate, because we focus on the use case of AutoML. AutoML frameworks, like Auto-Sklearn, do not employ any additional constraints. 
Nevertheless, we want to support additional constraints in the future. 
For example, some ensemble techniques work best with a diverse set of base models \citep{banfield2005ensemble}. 
Therefore, Assembled-OpenML with additional constraints could make interesting experimenters much cheaper computationally. 
For example, we could validate if it is a good idea to only store a diverse set of configurations during the run of an AutoML tool without having to expensively run AutoML tools\footnote{We explored this in preliminary experiments, see Appendix \ref{appdx:diverser} for more details.}. 

\section{Using Assembled-OpenML to Compare Ensemble Techniques}
To provide an example of how to use Assembled-OpenML to compare ensemble techniques, we created a simple benchmark set of metatasks (Section \ref{sec:createbenchmark}). Furthermore, we implemented a set of ensemble techniques such that we can simulate their behavior by passing prediction data to the technique (Section \ref{sec:simulate}).

\subsection{Creating a Benchmark using Assembled-OpenML}
\label{sec:createbenchmark}
For this example, we decided to use a list of OpenML task IDs from a curated benchmarking suite \citep{bischl2017openml} as input to Assembled-OpenML. 
We selected the benchmarking suite "OpenML-CC18"\footnote{See \url{https://www.openml.org/s/99} and \url{https://docs.openml.org/benchmark/}.}, which includes $72$ tasks. 
The tasks in OpenML-CC18 adhere to a list of criteria appropriate for a benchmark, such as having to use 10-fold cross-validation.
Thus, the first step is to run Assembled-OpenML for each task ID in OpenML-CC18. 
Here, we decided to use OpenML's Area Under ROC Curve (AUROC) metric to select the $50$ best performing configurations for each task (if more than $50$ exist). 
This took ${\sim} 55$ minutes without parallelization. 
The time it takes to build a metatask depends on the hardware, internet quality, and OpenML's response time\footnote{
All experiments in this paper are done on a workstation with an AMD Ryzen Threadripper PRO 3975WX CPU, SSD storage, 528 GB RAM, and a download speed of 1 Gbps. Moreover, we noticed no issues with OpenML's response time.}.

The resulting $72$ metatasks could already be used as a benchmark.
However, we also want to detail the possibility of post-processing a set of metatasks. 
To do so, we created a script that filters metatasks and base models based on the following constraints.

To quantify the potential of ensemble techniques for a dataset, we used concepts from Dynamic/Algorithm Selection \citep{cruz2018dynamic,DBLP:journals/ec/KerschkeHNT19}: the Virtual Best Algorithm (VBA) and the Single Best Algorithm (SBA). The VBA represents an oracle-like perfect selector that always selects the prediction of the best base model for an instance. The SBA represents the average best selector and returns the predictions of the base model that is assumed to be the best on average over all instances (e.g., has the highest score on the validation data). 
We use the difference in performance between VBA and SBA (called the \textit{VBA-SBA-Gap}) to filter metatasks. 
In other words, we assume for this benchmark that if a VBA-SBA-Gap exists, the task is more interesting for ensemble techniques. Having no gap is interpreted as one base model being as good as (naively) combining multiple base models. 
We required that metatasks have a VBA-SBA-Gap of $5\%$ in performance to guarantee that there is a (theoretical) room for improvement over the SBA. 
Moreover, we removed worse-than-random base models and filtered base models with corrupted prediction data.
Finally, we require that a valid metatask has at least 10 base models. 

This post-processing took ${\sim} 7$ minutes without parallelization. $31$ metatasks remained after post-processing. 
For details on these metatasks, refer to Table \ref{tab:finalmetatasks} in Appendix \ref{appdx:metatasktable}. 
We provide code to re-build the $31$ metatasks from an automatically created benchmark specification.

\subsection{Simulating Ensemble Techniques}
\label{sec:simulate}
We execute an ensemble technique without having to train and evaluate the base models. Thus, leaving only the computational overhead of the ensemble technique.
We deem this to be a simulation as any practical application of an ensemble technique would need to train and evaluate the base models. For example, scikit-learn's StackingClassifier\footnote{\url{https://scikit-learn.org/stable/modules/generated/sklearn.ensemble.StackingClassifier.html}} expects (untrained) base models as input. 

To simulate an ensemble technique, we use the data stored in a metatask. 
We also use the original task's folds from OpenML, because the prediction data was created using cross-validation.
For each fold, we split the fold's predictions on the test data in two halves with a ratio of $0.5$, creating meta-train and meta-test predictions. The split is done in a stratified fashion. 
Next, the meta-train predictions are used to train or build the ensemble technique. Lastly, the meta-test predictions are used to evaluate the ensemble technique.
As a result, we use a metatask to perform 10-fold cross-validation of the simulated ensemble techniques. 
Some ensemble techniques utilize the training data and not only the base models' predictions. We employed one-hot encoding and filled missing values with a static value to make the training data usable by such techniques.  

We simulated the following techniques: Stacking \citep{DBLP:journals/nn/Wolpert92}; majority Voting; ensemble selection from libraries of models \citep{caruana2004ensemble}; Dynamic Classifier Selection (DCS) \citep{giacinto2001dynamic}; Dynamic Ensemble Selection (DES) \citep{ko2008dynamic}; a SBA as well as a VBA versions of DCS (called DCS-SBA and DCS-VBA); and a novel oracle-like baseline called the Virtual Best Ensemble (VBE).
See Appendix \ref{appdx:ens_tech_details} for more details on the individual simulated ensemble techniques.

\subsection{Results and Summary}
\label{sec:results}
Obtaining the prediction data used in our comparison took $1$ hour and $2$ minutes, which is the combined time of sequentially fetching and building the $72$ original metatasks as well as reducing it to a benchmark set of $31$ metatasks. 
The total prediction data of the $31$ metatasks is equal to training and evaluating $1523$ base models. 
In comparison, obtaining the predictions by training and evaluating one base model, a Histogram-based Gradient Boosting Classification Tree\footnote{
Originally, we wanted to use the model with the highest AUROC for the runtime comparison. However, we were not able to initialize this model using OpenML's tools. We assume that this is a bug resulting from OpenML's development. See our code for more details. Consequently, we opted for the model with the next highest AUROC that we can initialize. 
},
on the most computationally expensive task/dataset, the dataset CIFAR\_10 with OpenML task ID $167124$, took ${\sim} 37$ minutes.
The training of the model used parallelization on all $64$ cores by default.
This excludes the time it took to find the model and to build an environment for its execution. 

We have no space for a detailed analysis of the example benchmark's results in this paper. Refer to Appendix \ref{appdx:result_figure} for more details on the results of running the simulated ensemble techniques on the benchmark created using Assembled-OpenML.
Simulating all ensemble techniques for all datasets across all $10$ folds took ${\sim} 4$ hours without parallelization. 

\section{Limitations and Broader Impact Statement}
\label{sec:limits}
As the limitations mentioned in Section \ref{sec:Assembled-OpenML} hint at, the biggest limitation of Assembled-OpenML is its data source. 
OpenML does not have enough data on, for example, regression tasks. 
Moreover, data stored on OpenML is sometimes problematic. We were not able to reproduce/initialize some base models\footnote{Here, we are referring to the model with the highest AUROC on the dataset CIFAR\_10 with OpenML task ID $167124$.}. Additionally, we found unexplainable problems with the prediction data\footnote{See Appendix \ref{appdx:parseproblems}.}.
Still, we believe that OpenML is the best publicly available data source. 

Assembled-OpenML enables less computationally expensive comparisons. Comparisons based on the data created by our tool could lead to re-evaluating the ensemble techniques used in existing (AutoML) applications.
Additionally, our work could lead to initiatives that try to produce less computational expensive benchmarks for ensembles. Thus, reducing costs and environmental impact. 
A negative impact of Assembled-OpenML could be an increase in traffic and cost of OpenML. We tried to minimize the API calls made by Assembled-OpenML. 
Generally, an initiative to share prediction data would also be helpful to make comparisons less computationally expensive. 

\section{Conclusion}
We presented the first version of Assembled-OpenML, a tool to generate metatasks using OpenML. Metatasks are meta-datasets that make it less computationally expensive to evaluate ensemble techniques. 
Additionally, we detailed an example of using Assembled-OpenML to compare ensemble techniques by building a benchmark set of metatasks and simulating ensemble techniques.

\newpage
\bibliography{references}

\newpage
\section{Reproducibility Checklist}

\begin{enumerate}
\item For all authors\dots
  \begin{enumerate}
  \item Do the main claims made in the abstract and introduction accurately
    reflect the paper's contributions and scope?
    \answerYes{The abstract and introduction both claim that we present Assembled-OpenML, an example benchmark, and an example of simulating ensemble techniques. We do exactly that in Section \ref{sec:Assembled-OpenML}, \ref{sec:createbenchmark}, and \ref{sec:simulate}.}
  \item Did you describe the limitations of your work?
    \answerYes{See Section~\ref{sec:limits}.}
  \item Did you discuss any potential negative societal impacts of your work?
    \answerYes{See Section~\ref{sec:limits}.}
  \item Have you read the ethics author's and review guidelines and ensured that your paper
    conforms to them? \url{https://automl.cc/ethics-accessibility/}
    \answerYes{We believe that our paper conforms to the guidelines.}
  \end{enumerate}
\item If you are including theoretical results\dots
  \begin{enumerate}
  \item Did you state the full set of assumptions of all theoretical results?
    \answerNA{We have no theoretical results.}
  \item Did you include complete proofs of all theoretical results?
    \answerNA{We have no theoretical results.}
  \end{enumerate}
\item If you ran experiments\dots
  \begin{enumerate}
  \item Did you include the code, data, and instructions needed to reproduce the
    main experimental results, including all requirements (e.g.,
    \texttt{requirements.txt} with explicit version), an instructive
    \texttt{README} with installation, and execution commands (either in the
    supplemental material or as a \textsc{url})?
    \answerYes{Code to re-create the data we have used exist in our GitHub repository\footnote{\url{https://github.com/ISG-Siegen/assembled}}. Moreover, we include all code used to generate results for this paper. We documented our work with a README and the dependencies with a requirements file.}
  \item Did you include the raw results of running the given instructions on the given code and data?
    \answerYes{The predictions of the simulated ensemble techniques for each metatask are stored in our GitHub repository.}
  \item Did you include scripts and commands that can be used to generate the
    figures and tables in your paper based on the raw results of the code, data,
    and instructions given?
    \answerYes{The evaluation folder of our GitHub repository contains the code used to generate our figures and tables.}
  \item Did you ensure sufficient code quality such that your code can be safely
    executed and the code is properly documented?
    \answerYes{We believe the code quality is sufficient. We include comments and annotations.}
  \item Did you specify all the training details (e.g., data splits,
    pre-processing, search spaces, fixed hyperparameter settings, and how they
    were chosen)?
    \answerYes{See Section~\ref{sec:simulate} and Appendix~\ref{appdx:ens_tech_details}. The details for the folds are omitted as they depend on OpenML and we did not create them.
    }
  \item Did you ensure that you compared different methods (including your own)
    exactly on the same benchmarks, including the same datasets, search space,
    code for training and hyperparameters for that code?
    \answerYes{All simulated ensemble techniques use the same code/data to be executed.}
  \item Did you run ablation studies to assess the impact of different
    components of your approach?
    \answerNA{We did not represent a new approach for which ablation studies are appropriate.}
  \item Did you use the same evaluation protocol for the methods being compared?
    \answerYes{See Section~\ref{sec:simulate}.}
  \item Did you compare performance over time?
    \answerNA{We do not have a time component in our experiments.}
  \item Did you perform multiple runs of your experiments and report random seeds?
    \answerYes{As a result of our data, we used 10-fold cross validation to evaluate the simulated techniques variability.
    However, we do not have the random seeds used to generate the splits. As far as we know, OpenML does not store the seeds used to create the splits.
    
    We have used a random state for our example benchmark to make our result reproducible across multiple executions. See our code for more details. However, we did not perform repetitions with different random states. 
    }
  \item Did you report error bars (e.g., with respect to the random seed after
    running experiments multiple times)?
    \answerYes{Our visualization of the results contains the performance on all folds. But the used visualization does not include the correct mapping from fold number to performance. This can be found in the raw results if needed.}
  \item Did you use tabular or surrogate benchmarks for in-depth evaluations?
    \answerNA{We do not have such benchmarks available. This is part of the motivation for our work.}
  \item Did you include the total amount of compute and the type of resources
    used (e.g., type of \textsc{gpu}s, internal cluster, or cloud provider)?
    \answerYes{See Section \ref{sec:createbenchmark}, Section \ref{sec:simulate} and \ref{sec:results}.}
  \item Did you report how you tuned hyperparameters, and what time and
    resources this required (if they were not automatically tuned by your AutoML
    method, e.g. in a \textsc{nas} approach; and also hyperparameters of your
    own method)?
    \answerNA{We did not tune hyperparameters. The simulated techniques are not sophisticated enough to include HPO yet. This is future work.}
  \end{enumerate}
\item If you are using existing assets (e.g., code, data, models) or
  curating/releasing new assets\dots
  \begin{enumerate}
  \item If your work uses existing assets, did you cite the creators?
    \answerYes{See Appendix~\ref{appdx:usedtools}.}
  \item Did you mention the license of the assets?
    \answerYes{See Appendix~\ref{appdx:usedtools}.}
  \item Did you include any new assets either in the supplemental material or as
    a \textsc{url}?
    \answerYes{Our tool Assembled-OpenML can be understood as a new asset. It is include per URL to our GitHub repository.}
  \item Did you discuss whether and how consent was obtained from people whose
    data you're using/curating?
    \answerYes{See Section \ref{sec:createbenchmark}, we are using OpenML-CC18 and its data. We cited all data sources according to the guidelines of datasets on OpenML (and in OpenML-CC18).}
  \item Did you discuss whether the data you are using/curating contains
    personally identifiable information or offensive content?
    \answerNA{Our data does not contain personally identifiable information or offensive content.}
  \end{enumerate}
\item If you used crowdsourcing or conducted research with human subjects\dots
  \begin{enumerate}
  \item Did you include the full text of instructions given to participants and
    screenshots, if applicable?
    \answerNA{We did not do research with human subjects.}
  \item Did you describe any potential participant risks, with links to
    Institutional Review Board (\textsc{irb}) approvals, if applicable?
    \answerNA{We did not do research with human subjects.}
  \item Did you include the estimated hourly wage paid to participants and the
    total amount spent on participant compensation?
    \answerNA{We did not do research with human subjects.}
  \end{enumerate}
\end{enumerate}

\appendix

\section{Details on Metatasks of the Example Benchmark}
\label{appdx:metatasktable}
\begin{table}[h]
    \centering
    \sisetup{zero-decimal-to-integer}
    \sisetup{round-mode=places}
    \caption{\textbf{Metatasks of the Example Benchmark} The table contains details on metatasks that were obtained by building the example benchmark. Each OpenML task is associated to a dataset. The $31$ tasks and datasets presented here were selected by post-processing/filtering the $72$ metatasks obtained by running Assembled-OpenML for all tasks of the curated benchmark suite "OpenML-CC18".}
    \label{tab:finalmetatasks}
    \resizebox{\textwidth}{!}{\begin{tabular}{l c | S[table-format=5, round-precision=0]  S[table-format=4.2, round-precision=2] S[table-format=2.2, round-precision=2] | S[table-format=2.2, round-precision=2]}
        \toprule
        Dataset & {Task ID} & {Instances} & {Features} & {Classes} & {Base Models}\\
        \midrule
        numerai28.6\footnotemark & 167120 & 96320 & 21 & 2 & 50\\
        connect-4 \citep{Dua:2019} & 146195 & 67557 & 42 & 3 & 50\\
        CIFAR\_10  \citep{krizhevsky2009learning} & 167124 & 60000 & 3072 & 10 & 50\\
        adult \citep{kohavi1996scaling,Dua:2019} & 7592 & 48842 & 14 & 2 & 50\\
        bank-marketing \citep{moro2014data} & 14965 & 45211 & 16 & 2 & 50\\
        jm1 \citep{Sayyad-Shirabad+Menzies:2005} & 3904 & 10885 & 21 & 2 & 50\\
        GesturePhaseSegmentationProcessed \citep{madeo2013gesture, Dua:2019} & 14969 & 9873 & 32 & 5 & 50\\
        first-order-theorem-proving \citep{bridge2014machine} & 9985 & 6118 & 51 & 6 & 50\\
        phoneme\footnotemark & 9952 & 5404 & 5 & 2 & 50\\
        Bioresponse\footnotemark & 9910 & 3751 & 1776 & 2 & 50\\
        madelon \citep{guyon2004result} & 9976 & 2600 & 500 & 2 & 50\\
        ozone-level-8hr \citep{zhang2008forecasting, Dua:2019}& 9978 & 2534 & 72 & 2 & 50\\
        kc1 \citep{Sayyad-Shirabad+Menzies:2005}& 3917 & 2109 & 21 & 2 & 50\\
        mfeat-morphological \citep{Dua:2019} & 18 & 2000 & 6 & 10 & 50\\
        steel-plates-fault \citep{provided2019research}& 146817 & 1941 & 27 & 7 & 50\\
        pc3 \citep{Sayyad-Shirabad+Menzies:2005}& 3903 & 1563 & 37 & 2 & 50\\
        cmc \citep{Dua:2019} & 23 & 1473 & 9 & 3 & 47\\
        pc4 \citep{Sayyad-Shirabad+Menzies:2005}& 3902 & 1458 & 37 & 2 & 50\\
        pc1 \citep{Sayyad-Shirabad+Menzies:2005}& 3918 & 1109 & 21 & 2 & 49\\
        qsar-biodeg \citep{mansouri2013quantitative} & 9957 & 1055 & 41 & 2 & 50\\
        credit-g \citep{Dua:2019} & 31 & 1000 & 20 & 2 & 50\\
        vehicle \citep{siebert1987vehicle} & 53 & 846 & 18 & 4 & 50\\
        analcatdata\_dmft \citep{simonoff2003analyzing} & 3560 & 797 & 4 & 6 & 50\\
        diabetes \citep{Dua:2019} & 37 & 768 & 8 & 2 & 50\\
        blood-transfusion-service-center \citep{yeh2009knowledge} & 10101 & 748 & 4 & 2 & 50\\
        eucalyptus \citep{bulloch1991eucalyptus} & 2079 & 736 & 19 & 5 & 50\\
        credit-approval \citep{Dua:2019} & 29 & 690 & 15 & 2 & 49\\
        ilpd \citep{Dua:2019} & 9971 & 583 & 10 & 2 & 50\\
        cylinder-bands \citep{evans1994overcoming,Dua:2019}& 14954 & 540 & 37 & 2 & 50\\
        climate-model-simulation-crashes \citep{lucas2013failure} & 146819 & 540 & 18 & 2 & 45\\
        kc2 \citep{Sayyad-Shirabad+Menzies:2005} & 3913 & 522 & 21 & 2 & 33\\
        \midrule
        MEAN & - & 12244.290323 & 193.387097 & 3.258065 & 49.129032\\
        \bottomrule
   \end{tabular}}
\end{table}
\addtocounter{footnote}{-2}
\footnotetext{\url{https://www.kaggle.com/datasets/numerai/encrypted-stock-market-data-from-numerai}} \addtocounter{footnote}{1}
\footnotetext{\url{https://sci2s.ugr.es/keel/dataset.php?cod=105}} \addtocounter{footnote}{1}
\footnotetext{\url{https://www.kaggle.com/competitions/bioresponse/data}}

\section{Performance Results of the Example Benchmark}
\label{appdx:result_figure}
In the following analysis, we want to determine the average best ensemble technique for post hoc ensembling. The example benchmark provides us with data that simulates the use case of an AutoML framework after model selection, whereby the $50$ best performing models have been pre-selected for ensembling.

In this use case, we are only interested in the performance w.r.t. OpenML's Area Under ROC Curve (AUROC) metric. That is, the metric that was used to select the $50$ best performing models. 
Post hoc ensembling is an extension of the optimization process within the AutoML tool. Hence, to evaluate an ensemble technique for post hoc ensembling, it must be analyzed w.r.t. the metric that is to be optimized during model selection. Extending this analysis to multiple metrics (and hence to multiple benchmarks) is left for future work. 

\begin{figure}[h!]
  \centering
  \includegraphics[height=0.8\textheight]{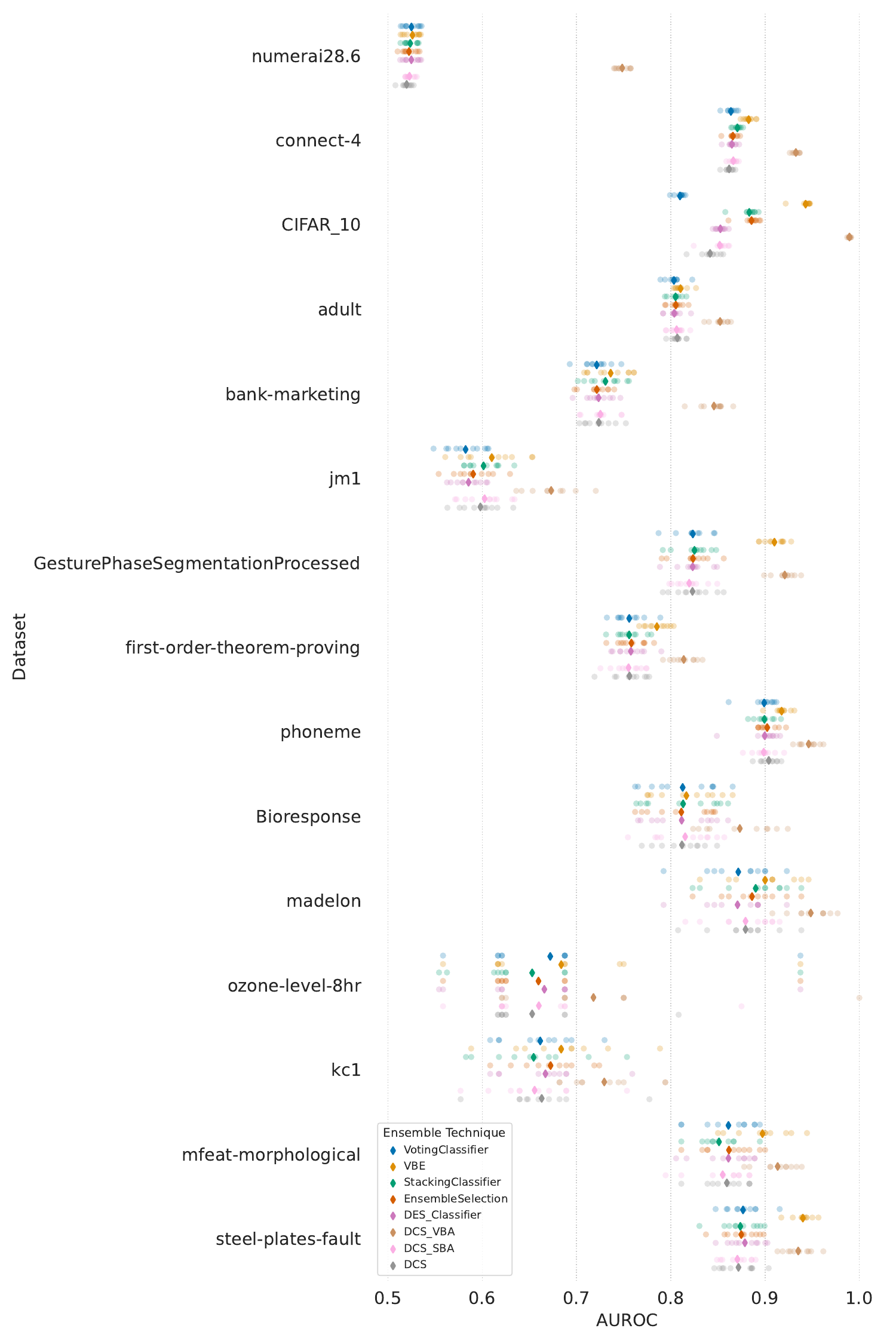}
  \caption{\textbf{Performance of Simulated Ensemble Techniques on the Example Benchmark - Part 1}
            The Area Under ROC Curve (AUROC) score of different ensemble techniques for all 10 folds per dataset and their mean (represented by the diamond). 
            }
    \label{fig:results}
\end{figure}

\begin{figure}[h!]
  \centering
  \includegraphics[height=0.8\textheight]{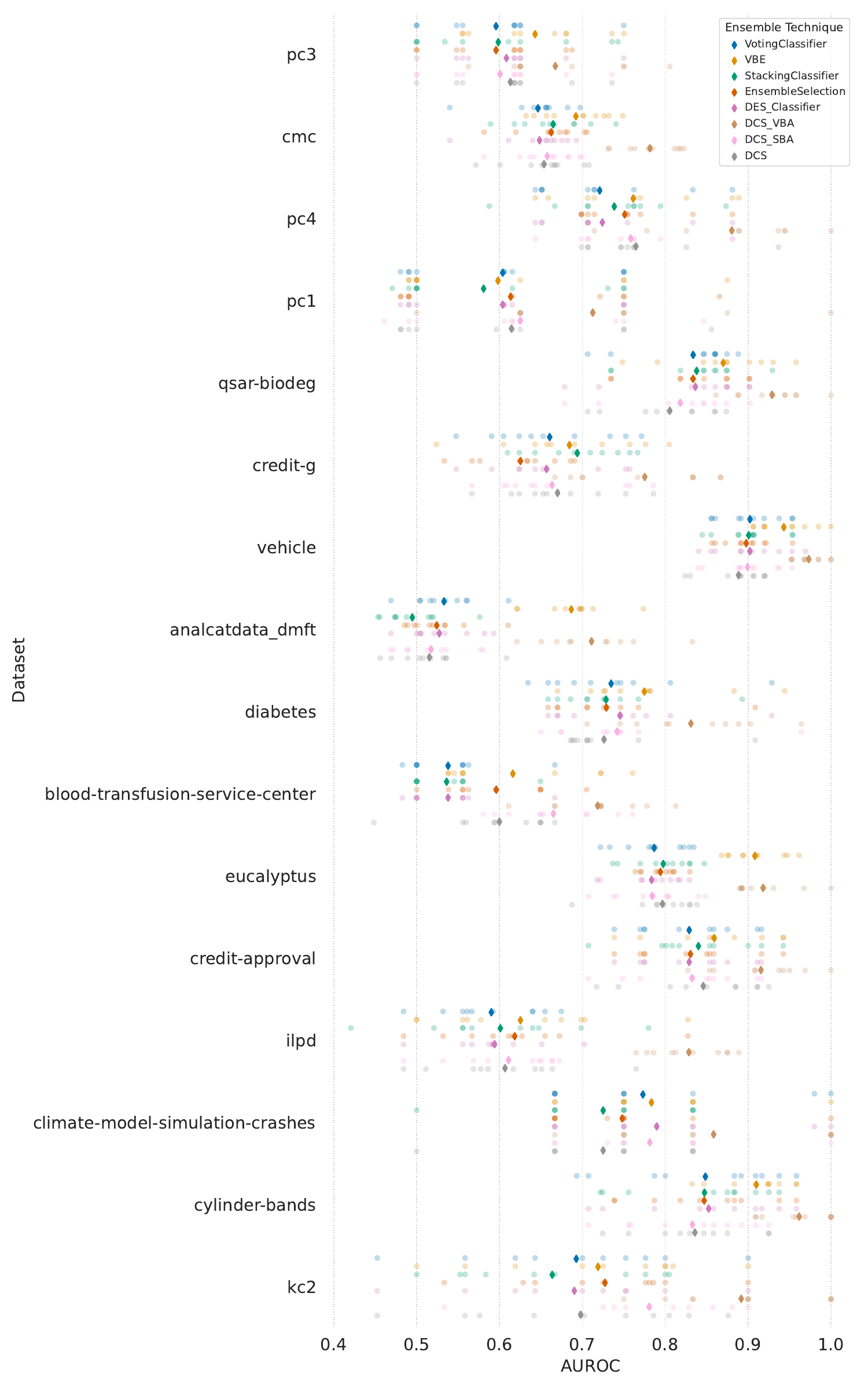}
  \caption{\textbf{Performance of Simulated Ensemble Techniques on the Example Benchmark - Part 2}
            The Area Under ROC Curve (AUROC) score of different ensemble techniques for all 10 folds per dataset and their mean (represented by the diamond). 
            }
    \label{fig:resultsp2}
\end{figure}

Please refer to Figure \ref{fig:results} and \ref{fig:resultsp2} for a detailed presentation of the performance of all ensemble techniques across all datasets. 
Both figures show that the performance can drastically differ per fold. This is a problem especially for smaller datasets. 
Considering that we use 10-fold cross-validation and split the fold's prediction data with a fraction of $0.5$, only $5\%$ of the data are used for training by the ensemble technique and another $5\%$ for evaluating the ensemble technique per fold. To illustrate, a dataset with 1000 instances would only have $50$ instances for training and $50$ more for evaluation per fold. 
Metatasks would require validation data to get more instances for the training phase of ensemble techniques. However, such data is not available on OpenML. 
As a result, evaluating with Assembled-OpenML might only be representative for larger datasets on OpenML.
We see it as future work to explore data sources that include validation data.


To determine the average best, we use the closed gap metric following \cite{DBLP:conf/oasc/LindauerRK17}. 
That is, we normalize the mean performance of each ensemble technique per dataset by using the mean performance of the VBA and SBA. 
We set the performance of the VBA equal to $1$ and the performance of the SBA equal to $0$. Any technique that has a higher performance than the SBA will get a positive value between $0$ and $1$ and a technique that performs worse than the SBA will get a negative value. The normalized value of an ensemble technique will show us how much it improved upon the SBA in relation to the VBA (or degraded performance for negative values). 
Finally, we take the mean over all datasets of all normalized values.
To overcome the impact of too small datasets, we additionally evaluate it once only for datasets with at least $1900$ samples in total. We select $1900$ as the threshold based on the dataset "steel-plates-fault" (OpenML Task ID $146817$) in Figure \ref{fig:results} and \ref{fig:resultsp2}. Datasets with more than $1900$ samples seem to vary less per fold performance. 
The results are shown in Table \ref{tab:normalized_results}. 

\begin{table}[h]
    \centering
    \caption{\textbf{Ensemble Technique Results} Shown are the mean and standard deviation in parentheses of the closed gap normalized performances for each ensemble technique on all datasets and only on datasets with at least $1900$ samples.
    The evaluated ensemble techniques are Dynamic Classifier Selection (DCS), Dynamic Ensemble Selection (DES), Stacking, Voting, Ensemble Selection (ES), Virtual Best Ensemble (VBE), and Virtual Best Algorithm (VBA). 
    }
    \label{tab:normalized_results}
    \resizebox{\textwidth}{!}{\begin{tabular}{l|l|lllll|ll}
        \toprule
        Data & SBA & DCS & DES & Stacking & Voting & ES & VBE &  VBA \\
        \midrule
        All Datasets  &  0.0 &  -0.09 (±0.29) &  -0.104 (±0.46) &    -0.063 (±0.28) &  -0.134 (±0.46) & -0.13 (±0.5) & 0.307 (±0.43)  &  1.0 (±0.0) \\
        Datasets $n \geq 1900$ &  0.0 &  -0.002 (±0.07)  &  0.005 (±0.1) & 0.023 (±0.09) &  -0.022 (±0.14) & 0.041 (±0.11) &  0.395 (±0.33)  &  1.0 (±0.0) \\
        \bottomrule
    \end{tabular}}
\end{table}

We can see from Table \ref{tab:normalized_results} that Ensemble selection and Stacking perform best.
The average best ensemble technique including small datasets is Stacking. Excluding small datasets, Ensemble Selection is the average best. 
For larger datasets, both are able to improve upon the SBA on average. Likewise, for all ensemble techniques, the standard deviation is decreased and the mean performance is increased without smaller datasets.
The high standard deviation and worse-than-SBA performance of all ensemble techniques while including small datasets further indicate that separate validation data per fold or larger datasets should be used to give the ensemble techniques enough data for training. 
In general, the subpar performance can also be explained by the missing diversity of base models in the benchmark. Selecting the top-n configurations can result in selecting models of only two different algorithms, e.g., for the OpenML Task $31$ the top $100$ configurations are from two different implementations of Random Forest\footnote{See \url{https://www.openml.org/search?type=task&sort=runs&id=31}}. This shows that more sophisticated methods to filter the configurations obtained from OpenML might be needed.

The performance of the VBE shows that the VBA is perhaps not the best oracle that we can use to compare ensemble techniques.
As discussed in \cite{caruana2004ensemble}, we would like to set the upper limit to the Bayes optimal performance for normalization if we knew it. 
The large gap between VBE and VBA could indicate that the VBA is not close to the Bayes optimal. 
Problems of the oracle for dynamic selection were already raised by \cite{cruz2018dynamic}. 

In this initial version, Assembled-OpenML generated and stored the (meta-)data used for the comparisons. However, the usability of the evaluation can be improved upon. For the evaluation, we had to pass the prediction data to our own implementations of the ensemble techniques. In the future, we aim to automate passing the required data to any ensemble technique and evaluating the technique across all folds. 

\newpage

\section{Details on Simulated Ensemble Techniques}
For this paper, we implemented the ensemble techniques ourselves.
In the future, we want to reuse existing ensemble technique implementations and pass our data to the implementations. Assembled-OpenML shall support this by passing the data in the form of simulated/faked base models. Initial work on this has already begun. 
For this paper, the following ensemble techniques have been implemented:

\label{appdx:ens_tech_details}
\begin{itemize}
\item \textbf{Stacking Classifier:} We simulated an implementation of stacked generalization \citep{DBLP:journals/nn/Wolpert92} (also called stacking).
Thereby, a meta-learner learns to combine the predictions of the base models to predict the groundtruth. We use a default\footnote{Default values from scikit-learn version 1.0.2.} LogisticRegression from scikit-learn \citep{scikit-learn} as a meta-learner. We increased the number of maximum iterations to $1000$. For our benchmark, we simulate stacking with the predictions' confidences. We are not using the so-called "passthrough" option, that is, training the meta-learner on the predictions and the original training data. 

\item \textbf{Voting Classifier:} We simulated a voting classifier (like sklearn's VotingClassifier). A voting classifier combines the predictions of base models through a majority voting rule. For our benchmark, we simulate voting with the predictions (i.e., "hard" voting). 

\item \textbf{Ensemble Selector} We simulate ensemble selection from libraries of models \citep{caruana2004ensemble}. We implemented an ensemble selector similar to Auto-Sklearn's ensemble technique \citep{feurer-neurips15a}. In ensemble selection, an ensemble is built in a greedy and iterative fashion such that the performance on a validation set is maximized by the non-weighted average of the selected base model's predictions. Thereby, base models can be selected multiple times. The frequency of selection is used as a weight for each base model's predictions at test time. We use an ensemble size (number of iterations) of $50$ inspired by Auto-Sklearn's default value. 

\item \textbf{Dynamic Classifier Selector} We simulated an implementation of Dynamic Classifier Selection (DCS) \citep{giacinto2001dynamic}. DCS tries to select the best classifier to classify each instance. This is related to \textit{per instance algorithm selection} \citep{rice1976algorithm}. For example, the classifier for a new instance can be selected based on the performance of the base models on (training) instances in the neighborhood of the new instance. For our benchmark, we simulate DCS using a default RandomForestRegressor from sklearn as an empirical performance model (EPM) of the prediction error \citep{DBLP:journals/ai/HutterXHL14}. We select the classifier for which the EPM predicts the lowest error. In this simulation, selecting a classifier means returning the classifier's predictions. 

\item \textbf{Dynamic Ensemble Selector}: We simulated an implementation of Dynamic Ensemble Selection (DES) \citep{ko2008dynamic}. For each new instance, DES selects a subset of classifiers on the fly for which the predictions are aggregated. Selection techniques similar to DCS can be used. For our benchmark, we extended our Dynamic Classifier Selector implementation to return the combined predictions of a subset of classifiers. We also use a RandomForestRegressor as an EPM. We select the subset of classifiers by adding classifiers to the subset until the accumulated predicted error of the subset is greater than 50\% of the total predicted error for an instance. We combine the predictions of a subset of classifiers using majority voting. 

\item \textbf{Dynamic Classifier Selector - SBA}: A simulated version of DCS that always returns the predictions of the best single classifier on the training data. In other words, this is the SBA for selection.

\item \textbf{Dynamic Classifier Selector - VBA}: A simulated version of DCS that always returns the predictions of the classifier selected by the oracle-like VBA. 

\item \textbf{Virtual Best Ensemble - VBE}: We introduce a novel baseline called Virtual Best Ensemble (VBE). The VBE is a non-real oracle-like predictor to represent the case where a weighting ensemble technique (like stacking or ensemble selection) found the optimal set of weights for the test data on the training data. For simplicity, we assume that learning the weights on the test data finds an optimal set of weights for the test data. Thus, we use our Stacking Classifier implementation trained on the test data.  

\end{itemize}

\section{Problems with the Prediction Data of OpenML}
\label{appdx:parseproblems}
Obtaining predictions of a run is currently not integrated into the OpenML Python API\footnote{However, initial work on this topic exists: \url{https://github.com/openml/openml-python/pull/1128}}.
We found that there are multiple prediction file formats, specifically different column names exist. 
While this is a minor problem, which we solved by checking all formats encountered so far, the bigger problem is a discrepancy between the predictions and the predictions' confidence values. 

We have found many examples where the prediction was not equal to the class with the highest confidence value.
This is an expected problem for some algorithms, for which the confidence values that are computed are not representative (for example, sklearn's SVM\footnote{\url{https://scikit-learn.org/stable/modules/generated/sklearn.svm.SVC.html}}). Moreover, we also sometimes expect this to be a problem of numerical precision. In both cases, we can fix the confidence values. 
However, we also found discrepancies that we cannot explain and, therefore, do not know how to fix them\footnote{For more details and an example, see: \url{https://github.com/openml/openml-python/issues/1131}}. 
To handle this, we store information on the discrepancy and make it possible to filter runs with unexplainable discrepancies later on. In our examples, we always filtered unexplainable discrepancies. 

\section{Fetching the Best Algorithms Instead of Configurations for More Diverse Base Models}
\label{appdx:diverser}
Ensembles can perform better with a diverse set of base models \citep{banfield2005ensemble}.
Yet, by fetching the top-n configurations, Assembled-OpenML does not necessarily provide a diverse set of base models. 
We explored the possibly of fetching a more diverse set of base models in preliminary experiments. 
To do so, we tried fetching the best configuration of the top-n best performing algorithms/pipelines (called flows on OpenML) for a task instead of fetching the overall top-n best performing configurations. 
This can also be understood as an algorithm selection use case.

Although we can use the configurations of the top-n flows to produce a much more diverse set of base models, it drastically reduces the amount of usable data on OpenML. 
In total, only ${\sim}1600$ flows exist compared to millions of runs.
Furthermore, ensuring diversity in a collection of flows is problematic due to duplicated algorithms/pipelines. See Appendix \ref{appdx:openmlflowduplciates} for more details on the duplicate problem of OpenML flows.
Finally, we abandoned using the configurations of the top-n flows as base models because it seems to be too far away from the AutoML use case.

\section{Complications with OpenML Flow Duplicates}
\label{appdx:openmlflowduplciates}
An OpenML flow can capture any algorithm/pipeline of supported ML frameworks.
Yet, determining that two flows are duplicates of each other is complicated. 
OpenML stores the results of multiple ML frameworks. 
Hence, to ensure no duplicates across ML frameworks, we had to identify similar algorithms while being named differently across frameworks. To illustrate, scikit-learn calls a Random Forest Classifier "RandomForestClassifier" while mlr3 calls it "mlr.ranger". The two different implementations of the same algorithm are named differently. While a difference in implementation can affect performance, the difference might not be substantial enough.
The two implementations might not be different enough to not be deemed duplicates for the sake of diversity in base models.   
Moreover, OpenML often stores multiple versions of algorithms. Hence, we had to remove such version duplicates.
Lastly, minor changes in a pipeline can already create a new flow object stored on OpenML. 
For example, we found minor differences such as "$sklearn.imputer$" vs. "$sklearn\_.imputer$". 
To automatically associate flows with such minor changes to each other, a sophisticated analysis of the flow object is needed. We solved this by comparing the similarity of two flows manually\footnote{To not be overwhelmed with too much manual comparison effort, we only manually check two flows if they appear to be (highly) similar based on the edit distance.}. 

\section{Python Libraries used by Assembled-OpenML}
\label{appdx:usedtools}
The following Python libraries are used by Assembled-OpenML in its current version:
\begin{itemize}
    \item OpenML-Python \citep{feurer-arxiv19a}, Version 0.12.2, BSD 3-Clause License;
    \item Pandas \citep{reback2020pandas}, Version 1.4.1, BSD 3-Clause License;
    \item Requests\footnote{\url{https://pypi.org/project/requests/}}, Version 2.27.1, Apache License 2.0;
    \item SciPy \citep{2020SciPy-NMeth}, Version 1.8.0, BSD 3-Clause License;
    \item NumPy \citep{harris2020array}, Version 1.22.3, BSD 3-Clause License;
    \item python-Levenshtein\footnote{\url{https://pypi.org/project/python-Levenshtein/}}, Version 0.12.2,  GPL-2.0 License.

\end{itemize}

\end{document}